%% file: acl_latex.tex
\documentclass[11pt]{article}

\usepackage{acl}

\usepackage{booktabs}
\usepackage{colortbl}
\usepackage[table]{xcolor}
\usepackage{caption}   
\usepackage{pgf}       

\input{defns}

\usepackage{times}
\usepackage{latexsym}

\usepackage[T1]{fontenc}

\usepackage[utf8]{inputenc}

\usepackage{microtype}

\usepackage{inconsolata}

\usepackage{graphicx}

\usepackage{float}

\usepackage{subcaption}
\usepackage{placeins}
\usepackage{afterpage}

\usepackage{booktabs}
\usepackage[table]{xcolor}
\usepackage{multirow}
\usepackage{caption}
\usepackage{amsmath} 
\usepackage{bm} 

\title{Evaluating the Effect of Linguistic Relatedness on Cross-Lingual Transfer in Large Multilingual Automatic Speech Recognition}

\author{
  Andrei Florian$^{1}$ \quad
  Cynthia Jayne Amol$^{2}$ \quad
  Hope Kerubo Ombaba$^{2}$ \quad
  Xiaoyu Cui$^{1}$ \quad
  Boniface Mwau$^{2}$ \\
  \textbf{Biatus Maina Kamau$^{2}$ \quad
  Lilian Diana Awuor Wanzare$^{2}$ \quad
  Christiane Fellbaum$^{1}$ \quad
  Happy Buzaaba$^{1}$} \\[0.5em]
  $^{1}$Princeton University \quad
  $^{2}$Maseno University \\[0.3em]
  \texttt{andrei.florian@princeton.edu, happy.buzaaba@princeton.edu}
}

\begin{document}
\maketitle




\begin{abstract}

Extending automatic speech recognition (ASR) to low-resource African languages is constrained by the prohibitive demands of data collection at scale. A promising direction is to leverage the linguistic relatedness between a low-resource target language and languages previously seen by a model to reduce the volume of target-language data needed for effective adaptation. Although this approach has proven reliable for text-based models, its effectiveness in the speech domain remains contested. We employ a systematic controlled experimental design spanning six factors, two Africa-centric corpora, and four large ASR models, sequentially adapting on a related auxiliary language followed by the target to isolate whether linguistic relatedness reliably predicts cross-lingual transfer gains across these conditions. In every setting, pre-adaptation on related auxiliary languages yields no practically meaningful improvements once as little as one hour of target-language data is available, suggesting that relatedness alone may not reliably predict transfer gains in large multilingual ASR, or constitute an effective strategy for extending such models to low-resource languages.

\end{abstract}


\section{Introduction}


\begin{figure}[h]
  \centering
  \includegraphics[width=\linewidth]{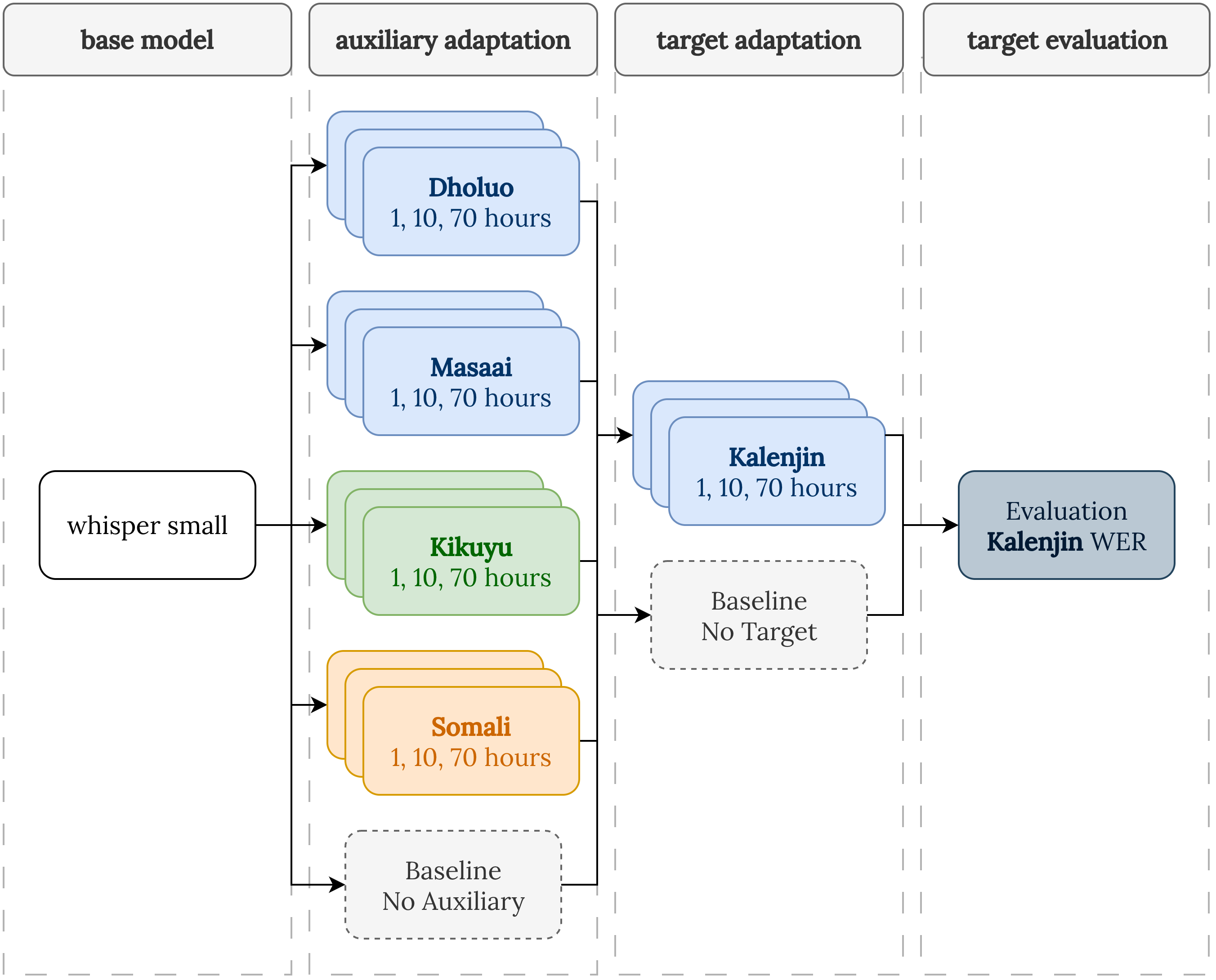}
  \caption{Experimental design for the first factor using languages from the AfriVoices KE corpus \citep{wanzare2026afrivoices}. Instances of the base Whisper Small \citep{radford2022robust} model are fine-tuned on each of four auxiliary languages, two linguistically related and two unrelated, at 1, 10, and 70 hours of labelled speech. The resulting auxiliary models, alongside the original baseline, are subsequently fine-tuned on 1, 10, and 70 hours of the target language Kalenjin, and evaluated using the Word Error Rate (WER) metric on the Kalenjin test set. Colours denote language family membership: blue (Nilotic), green (Bantu), orange (Cushitic).}
  \label{fig:fig1}
\end{figure}


Automatic Speech Recognition (ASR) holds particular promise for extending language technologies to communities whose languages are primarily oral, as is the case for the majority of Africa's approximately 2,000 languages \cite{orife2020masakhane}. Yet African languages remain profoundly under-represented in contemporary NLP systems. \citet{joshi2020state} estimate that 88\% of the world's languages have no meaningful presence in the datasets underpinning modern language models, a category encompassing most languages spoken across the African continent, and although recent large-scale models such as OpenAI's Whisper \cite{radford2022robust} and Meta's Omnilingual ASR \cite{keren2025omnilingual} have extended ASR capabilities across increasingly larger language sets, the vast majority of African languages remain systematically absent from their training corpora, leading to poor downstream performance.

The root of this deficit lies in the nature of the data collection problem. Many African languages are primarily oral, with limited written materials and a smaller fraction of those digitally available \cite{mbogho2025building}. As such, the web-scraping paradigms that enable large-scale corpus construction for high-resource languages do not transfer to this setting. Instead, building corpora for low-resource African languages typically requires manual elicitation from local speakers, followed by transcription and linguistic annotation \cite{nekoto2020participatory}. At the scale demanded by modern ASR training, this process is neither financially nor logistically viable for extending ASR models to the hundreds of African languages lacking digital resources \citep{mbogho2025building, nekoto2020participatory}.

A promising direction is to leverage cross-lingual transfer --- the tendency of multilingual models to form language-agnostic internal representations that allow knowledge acquired in one language to generalise to others --- to reduce the volume of target-language data needed for effective adaptation. However, whether linguistic relatedness reliably predicts transfer in speech-based models remains contested. We embed the two-stage fine-tuning methodology of \citet{pillai2024multistage} --- demonstrated to improve Whisper's performance on Malasar by sequentially adapting on the genealogically related auxiliary language Tamil followed by the target --- within a systematic controlled experimental design (Figure \ref{fig:fig1}) spanning six factors, two Africa-centric corpora, and four large ASR models of differing scale and architecture, isolating whether linguistic relatedness constitutes a reliable predictor of cross-lingual transfer effects across this broad range of conditions. In doing so, we assess the feasibility of relatedness-guided adaptation as a strategy for extending ASR capabilities to low-resource African languages.\footnote{Code available under GPL-3.0 at \url{https://anonymous.4open.science/r/salma2026-africanwhisper/}.}


\section{Related Work}

In multilingual text-based models, \citet{conneau2020unsupervised} and \citet{artetxe2020cross} demonstrate the emergence of shared representational subspaces across languages. \citet{radford2022robust} note that Whisper similarly learns cross-lingual representations through joint multilingual training, and \citet{babu2021xlsr} demonstrate the same capacity in wav2vec 2.0 models, suggesting that this capacity extends to multilingual ASR models. \citet{artetxe2020cross} and \citet{conneau2021unsupervised} further find, in text and speech models respectively, that these cross-lingual representations are especially well-aligned among linguistically related languages, which often share genealogy and syntactic, morphological, and phonological structure. \citet{lin2019choosing} and \citet{eronen2023zero} show that genealogical and typological proximity to previously seen languages predicts zero-shot cross-lingual transfer performance in text-based models, while \citet{lauscher2020zero} and \citet{wu2020all} demonstrate that such proximity enables target-language adaptation with substantially less data than would otherwise be required.

However, findings for speech-based models are less consistent and narrower in scope. \citet{san2024predicting} and \citet{rouditchenko2023comparison} find linguistic relatedness predictive of cross-lingual transfer in limited settings: the former showing that supplementing a low-resource target with a related donor language during pre-training improves target-language performance in XLS-R, and the latter that transfer to unseen languages is enhanced by the coverage of their language families in XLS-R and Whisper pre-training. Conversely, \citet{khare21_interspeech} and \citet{liebl2026aspects} find relatedness neither necessary nor consistently predictive: the former showing that pre-adapting wav2vec 2.0 on an unrelated auxiliary language still improves transfer to the target, and the latter that fine-tuning XLS-R on languages genealogically or phonologically similar to an unseen target does not reliably improve zero-shot transfer to that target across the three languages tested. These conflicting findings, arising from disparate experimental settings, motivate a controlled evaluation of the effect of relatedness on cross-lingual transfer in speech-based models.


\begin{figure*}[t]
  \centering
  \includegraphics[width=\linewidth]{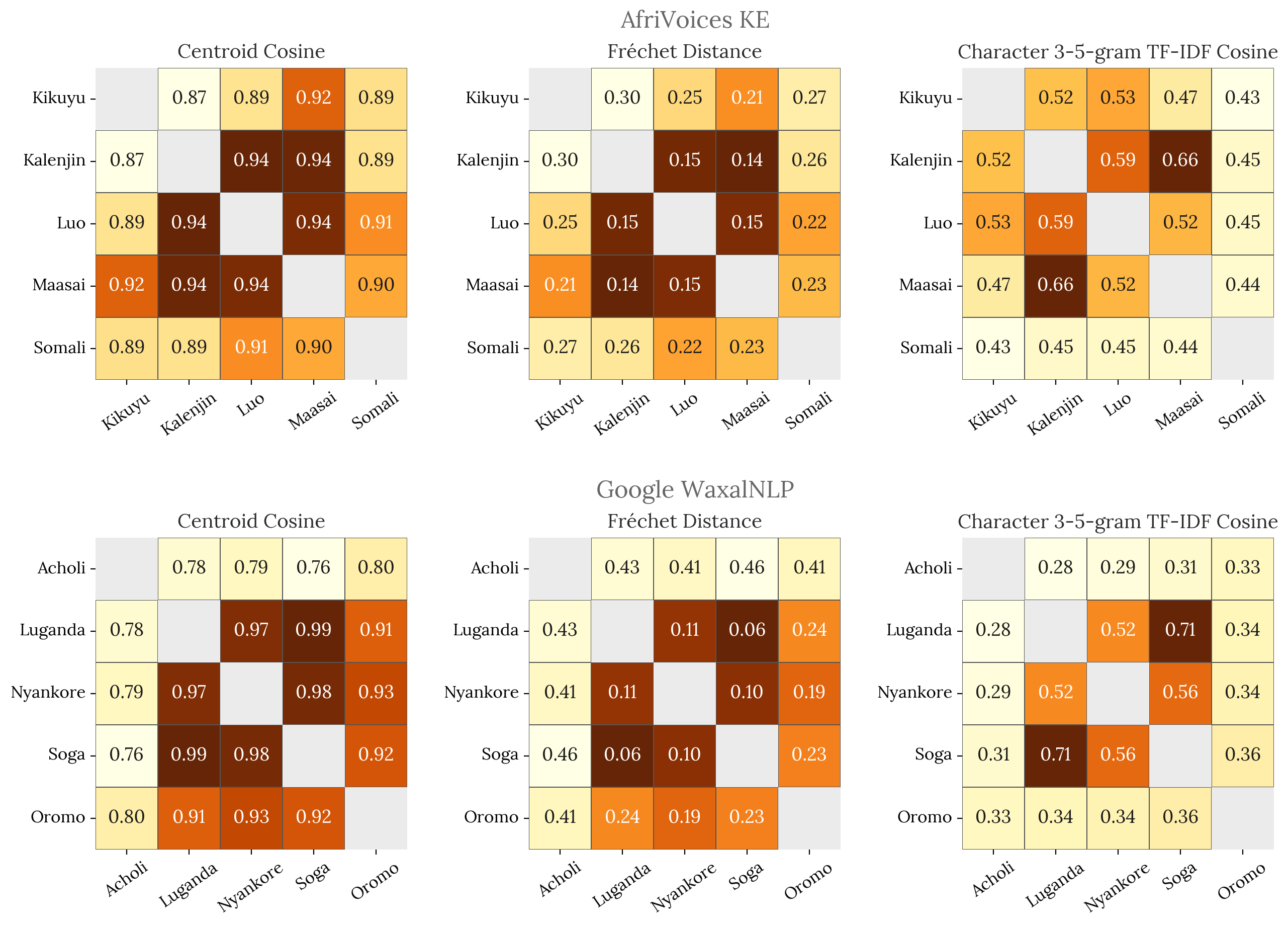}
  \caption{Corpus-level pairwise similarity results across three metrics for all languages in the AfriVoices KE (top) and Google WaxalNLP (bottom) corpora. Centroid cosine similarity is uniformly high across most language pairs in both corpora. Fréchet distance and character 3--5-gram TF-IDF cosine similarity both exhibit clear within-family clustering across both corpora.}
  \label{fig:fig2}
\end{figure*}


\section{Language Corpora and Models}

We draw on five languages from each of two multilingual, Africa-centric ASR corpora: AfriVoices KE \cite{wanzare2026afrivoices} and Google WaxalNLP \cite{diack2026waxal}.\footnote{AfriVoices KE is released under CC BY-SA 4.0 and Google WaxalNLP under CC BY-SA 4.0 and CC-BY-SA 4.0.} Each set of five languages spans three families --- Nilo-Saharan Nilotic, Niger-Congo Bantu, and Afro-Asiatic Cushitic --- spoken across Central and East Africa. Both corpora were compiled through analogous processes in which scripted and free-speech recordings from recruited native speakers were transcribed and annotated by teams of linguists \cite{wanzare2026afrivoices, diack2026waxal}. Languages are selected to reflect meaningful typological contrast between within-family and cross-family pairings across syntactic, morphological, and phonological dimensions, as characterised in the subsections below. This ensures that family membership tracks multi-dimensional linguistic similarity, constituting a principled proxy for relatedness in the experimental design. Full dataset statistics for all ten languages across both corpora are provided in Table~\ref{tab:dataset_stats}.

\subsection{Typological Overview}

Across the languages from the AfriVoices KE corpus \cite{wanzare2026afrivoices}, Dholuo, Maasai, and Kalenjin belong to the Nilotic family, sharing agglutinative morphology and lexical tone. Furthermore, Dholuo and Kalenjin show Subject-Verb-Object (SVO) word order, which is one point of distinction from the Verb-Subject-Object (VSO) order of Maasai. Although Kikuyu, a Bantu language, shares lexical tone and agglutinative morphology with the Nilotic family, it is distinct from this group through morphemes indicating noun class membership being prefixed to the root rather than added as suffixes to the end. Somali, a Cushitic language, also differs from the Nilotic set in that it has Subject-Object-Verb (SOV) syntax, relies on pitch accent, rather than tone, to distinguish word meanings, and its sound inventory includes phonemes found in Semitic languages.

The Google WaxalNLP corpus \cite{diack2026waxal} spans 27 languages, of which we select five representing the same three families. From the Bantu family, we choose Luganda, Nyankole, and Soga, which are agglutinative, show lexical tone, SVO order, and prefixed noun class morphemes. Acholi represents the Nilotic language family in this corpus and similarly has SVO syntax, although differing from the Bantu set in its morphosyntactic properties and complex tonal system. The Cushitic family is represented by Oromo, which is most distinct from the Bantu and the Nilotic languages through its SOV order and use of pitch accent.

The patterns described above suggest that these syntactic, morphological, and phonological characteristics cluster with family membership, though the low-resource languages considered here lack the extensive linguistic documentation required for a fully rigorous account. The corpus-level pairwise similarity analyses that follow complement them by providing quantitative proxies for typological contrast across multiple dimensions.

\subsection{Corpus-Level Analysis}

We perform three pairwise similarity analyses across all languages in both corpora to assess cross-corpus comparability, and further validate the language groupings underpinning the experimental design. Prior to the analysis, we remove recordings longer than 30 seconds, annotation tags, and utterances with fewer than 3 tokens or greater than the 99th-percentile length. 

We focus on transcript-based measures because the ten languages use largely phonemic orthographies at the segmental level, making textual similarity a reasonable proxy for phonological and morphological similarity. Several languages in both corpora employ lexical tone or pitch accent as phonologically distinctive features not captured by the transcriptions, though the within-family clustering we observe remains interpretable as reflecting genuine segmental proximity. Audio-based measures were additionally considered, but pitch and formant features proved insufficiently stable across corpora to support reliable comparison.

Centroid cosine similarity encodes transcripts using the paraphrase-multilingual-MiniLM-L12-v2 \citep{reimers2019sbert} sentence transformer and computes cosine similarity between per-language mean embeddings. We use this primarily as a corpus-comparability check, since high centroid similarity indicates topically comparable corpora. Fréchet distance compares both the mean and covariance of per-language transcript embedding distributions \citep{heusel2017gans, pillutla2021mauve}, capturing distributional shape and dispersion beyond mere topic proximity, which we interpret as a probe of morphosyntactic and lexico-semantic similarity. Character n-gram TF-IDF cosine similarity captures subword and surface-form patterns, providing a proxy for orthographic, lexical, morphological, and, to a limited extent, phonological overlap, characteristics particularly informative for the morphologically rich, agglutinative languages considered \citep{bojanowski2017enriching}.

As shown in Figure \ref{fig:fig2}, absolute centroid cosine similarity is uniformly high across most language pairs in both corpora, indicating broad comparability in topic and elicitation structure. This reduces the risk that transfer effects are influenced by corpus-topic mismatch. By contrast, Fréchet distance and character n-gram TF-IDF similarity exhibit clear within-family clustering: in AfriVoices KE, the Nilotic languages are consistently closer to one another than to Kikuyu or Somali, while in WaxalNLP, Luganda, Nyankole, and Soga cluster more tightly together than with Acholi or Oromo.

Together, these analyses indicate cross-corpus topical comparability and corroborate typological contrast between within-family and cross-family pairings across textual and embedding-based measures, lending validity to the family-based language groupings as a principled proxy for multidimensional relatedness in the experimental design.

\subsection{Models}

We extend the sequential fine-tuning setup of \citet{pillai2024multistage} to four large ASR models spanning a range of parameter scales, supervision paradigms, and architectural families, with OpenAI Whisper Small \cite{radford2022robust} as the experimental base. Whisper Small is a 244-million parameter encoder–decoder Transformer trained on 680,000 hours of weakly supervised multilingual audio–text pairs across 99 languages, noted for its effectiveness in low-resource adaptation \cite{radford2022robust} and fine-tuning efficiency due to its relatively modest size. Whisper Large v3 \cite{radford2022robust} scales the same architecture to 1,550 million parameters across 32 encoder and decoder layers, trained on 5 million hours of weakly and pseudo-labelled audio. This model pairing allows us to isolate the effect of model scale on cross-lingual transfer while controlling for architecture and supervision regime.

Facebook XLS-R \cite{babu2021xlsr} and Facebook Omnilingual ASR \cite{keren2025omnilingual} further vary both architecture and supervision regime, employing a self-supervised encoder-only wav2vec 2.0 backbone decoded via CTC rather than an autoregressive encoder–decoder. XLS-R comprises 300 million parameters trained on 436,000 hours of unlabelled speech across 128 languages, which OmniASR scales to 6.5 billion parameters and 4.3 million hours spanning around 1,600 languages.

All four models maintain a single unified parameter set shared across languages, permitting fine-tuning without architectural modification. Of the languages they are fine-tuned on in our experiments, both Whisper and XLS-R models have pre-training coverage of exclusively Somali, while OmniASR has pre-exposure to all AfriVoices KE languages except for Maasai \cite{radford2022robust, babu2021xlsr, keren2025omnilingual}.


\section{Methods}

We adopt the two-stage sequential fine-tuning methodology of \cite{pillai2024multistage}, applying it through a systematic controlled experimental design spanning six factors to isolate whether linguistic relatedness between the auxiliary and target reliably predicts cross-lingual transfer gains in these settings. We begin by filtering out all recordings exceeding 30 seconds in length from languages across both the AfriVoices KE and WaxalNLP corpora, as this is the maximum input length supported by Whisper models \cite{radford2022robust}. All remaining data is partitioned into train (85\%), test (10\%), and validation (5\%) sets for each language.

\textbf{First Factor.} We sample 1, 10, and 70 hours from the train set of each auxiliary language in the AfriVoices KE corpus, with 70 hours as the upper bound determined by the smallest per-language training set after filtering. Dholuo and Maasai constitute the related auxiliary set, Kikuyu and Somali the unrelated auxiliaries, and Kalenjin the target language. In the first stage, we full-model fine-tune a separate Whisper Small instance on each combination of auxiliary language and data volume, yielding 12 auxiliary models as shown in Figure \ref{fig:fig1}. In the second stage, each auxiliary model and the unmodified Whisper Small baseline are further fine-tuned on 1, 10, and 70 hours of Kalenjin, producing 39 target models. All auxiliary and target models are evaluated on the Kalenjin test set.

\textbf{Second Factor.} We repeat the setup of the first factor, replacing full-model fine-tuning with parameter-efficient LoRA for both fine-tuning stages, to assess whether our results remain consistent across varying fine-tuning methodologies.

\textbf{Third Factor.} We follow the first factor setup with the roles of Dholuo and Kalenjin exchanged, making Dholuo the target language and Kalenjin an auxiliary, to assess whether our results hold under a different within-family auxiliary-target pairing.

\textbf{Fourth Factor.} Mirroring the setup of the first factor, we introduce an intermediate fine-tuning stage, adapting the model on two related auxiliary languages --- Dholuo followed by Maasai --- prior to target-language fine-tuning on Kalenjin, to assess whether pre-adaptation on multiple related auxiliary languages affects cross-lingual transfer.

\textbf{Fifth Factor.} We sample 1, 10, and 25 hours from the train split of each language in the WaxalNLP corpus and repeat the first factor setup with Nyankole and Soga (Bantu) constituting the related auxiliary set, Acholi (Nilotic) and Oromo (Cushitic) the unrelated auxiliaries, and Luganda the target language, to assess whether our results generalise across corpora and language families.

\textbf{Sixth Factor.} We reproduce the first factor setup across three additional models --- Whisper Large v3, XLS-R, and OmniASR --- to assess whether our findings generalise across large ASR models of varying scale, architecture, and supervision.

Across all factors, we report Word Error Rate (WER) as the primary evaluation metric for all auxiliary and target models. We compare the corpus-level WER (the length-weighted mean of per-utterance WER computed case-insensitively and with punctuation removed, as reported in Table \ref{tab:wer_all_factors}) of models pre-adapted on related and unrelated auxiliary languages at each target-data volume across all factors, testing whether linguistic relatedness between the auxiliary and target improves transfer gains over unrelated pre-adaptation. We conduct an additional analysis, comparing corpus-level WER between the group of models pre-adapted on a related language and the target-only baseline, testing whether pre-adaptation on a related language improves target performance over direct target-only fine-tuning. Together, these analyses isolate whether any observed transfer gains are directly attributable to linguistic relatedness. 

For each analysis, we estimate the difference in corpus-level WER between groups and derive a $90\%$ confidence interval on this difference from a non-parametric bootstrap over test utterances ($5{,}000$ resamples). We take $\Delta = \pm 5$ absolute WER percentage points as our equivalence bounds, a performance difference beyond which we consider practical utility to be materially affected. We classify each comparison by where its confidence interval sits relative to the bounds: \textit{practically equivalent} when it falls entirely within them, \textit{practically meaningful} when it lies wholly beyond a bound, and \textit{inconclusive} when it straddles one. This equivalence criterion is inherited from a two one-sided tests (TOST) procedure \cite{lakens2017equivalence} at $\alpha = 0.05$ (full statistical results in Tables~\ref{tab:appendix_claim_a}, \ref{tab:appendix_claim_b}).

\textbf{Full-model fine-tuning.} All models are fine-tuned on a single NVIDIA H200, with a total computational budget of 1,020 GPU hours. We use a cosine learning-rate schedule, 0.01 weight decay, and mixed precision (bf16/TF32). Hyperparameters are tuned via a small grid over learning rate and number of epochs, with settings scaled to data volume. The final checkpoint is selected by minimum validation WER under greedy decoding, using early stopping with a patience of four evaluations. The Whisper models are fine-tuned using HuggingFace's \textit{Seq2SeqTrainer} with an effective batch size of 16 and gradient checkpointing enabled. XLS-R is fine-tuned using the standard \textit{Trainer} with an effective batch size of 32 and gradient checkpointing disabled, given its smaller memory footprint at 300M parameters. OmniASR is fine-tuned via a custom \textit{CTCTrainer} with an effective batch size of 16 and gradient checkpointing disabled, as the \textit{fairseq2} interface does not expose it. The Whisper and OmniASR models use 8-bit AdamW to reduce optimiser state memory and XLS-R uses fused AdamW.

\textbf{LoRA fine-tuning.} LoRA fine-tuning is only applied to Whisper Small in the second factor. We apply standard flat LoRA \cite{hu2022lora} to five module types: \textit{q\_proj}, \textit{v\_proj}, \textit{out\_proj}, \textit{fc1}, and \textit{fc2}, targeting both attention and feed-forward sublayers following evidence that FFN modules receive disproportionately high importance scores during low-resource cross-lingual ASR adaptation \citep{zhang2023adalora, song2024lora}. The first three encoder layers are frozen throughout, with adapters applied to encoder layers 3–11 and all decoder layers \cite{liu2024exploration}. Rank and learning rate scale with data volume, taking $(4, 10^{-4})$, $(8, 10^{-4})$, and $(8, 7{\times}10^{-5})$ at 1, 10, and 70\,h respectively; alpha is fixed at 16 and dropout at 0.05.


\section{Results}


\begin{figure*}[h]
    \centering
    \begin{subfigure}[t]{0.49\textwidth}
        \centering
        \includegraphics[width=\textwidth]{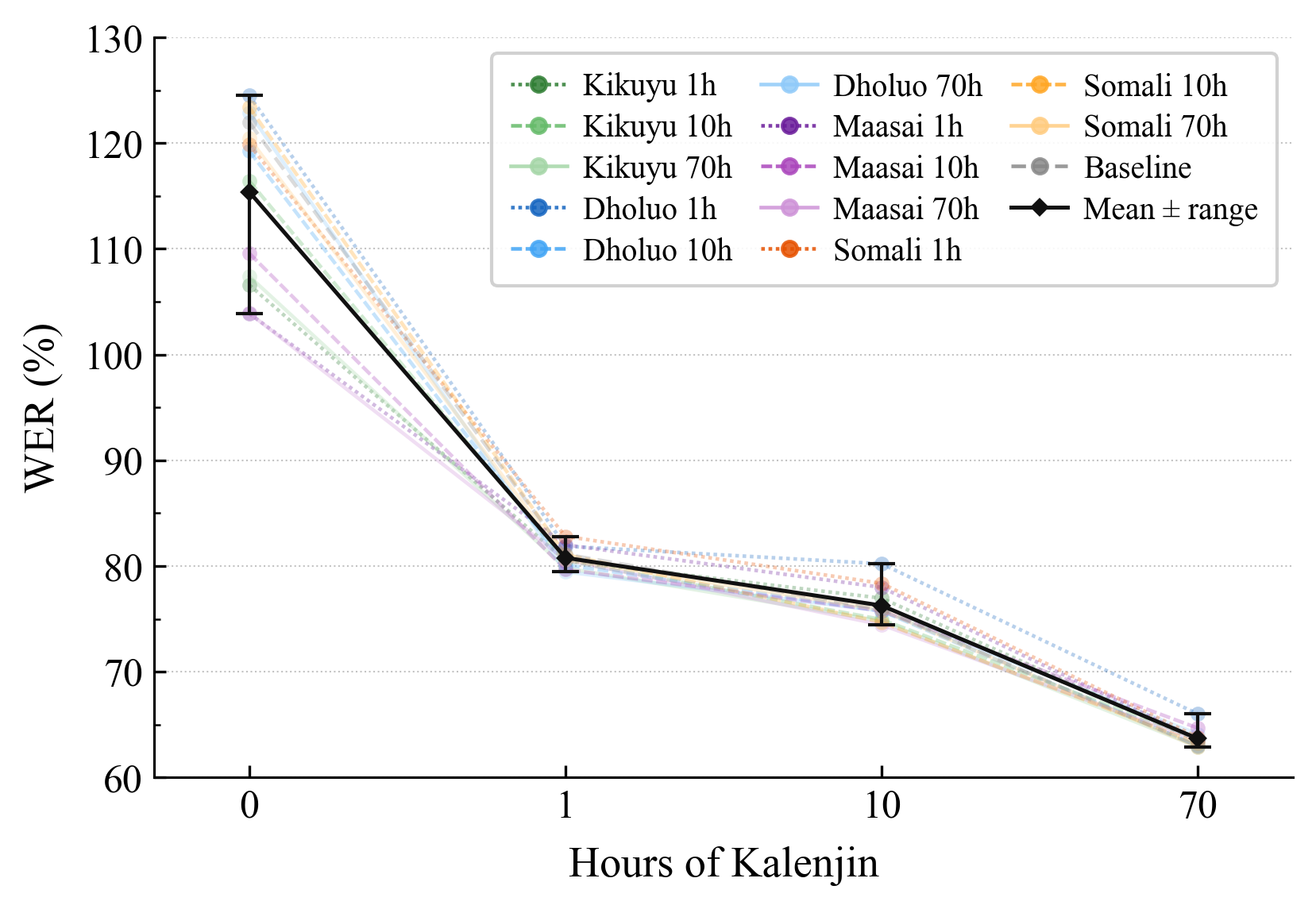}
        \label{fig:factor1_aux}
    \end{subfigure}
    \hfill
    \begin{subfigure}[t]{0.49\textwidth}
        \centering
        \includegraphics[width=\textwidth]{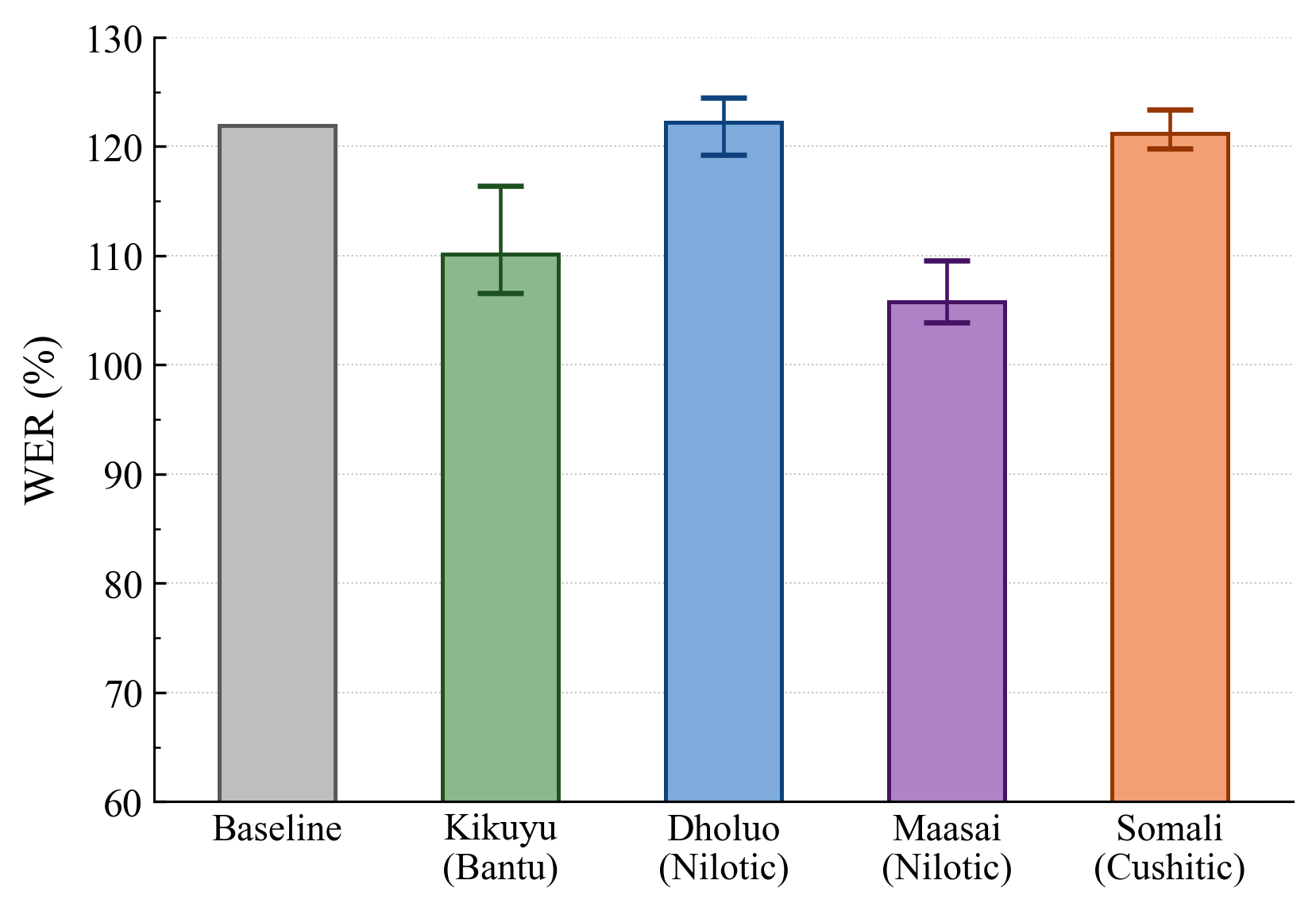}
        \label{fig:factor1_aux}
    \end{subfigure}
    \caption{WER on the Kalenjin test set across all first-factor models. \textit{Left}: target-language WER as a function of Kalenjin fine-tuning hours for all 12 auxiliary models and the unmodified baseline. The mean WER and range across auxiliary models at each volume of Kalenjin adaptation are marked in black. \textit{Right}: Mean WER of each auxiliary model on the Kalenjin test set prior to second-stage fine-tuning, with error bars denoting the range across auxiliary language data volumes.}
    \label{fig:fig3}
\end{figure*}


\begin{figure*}[h!]
    \centering
    \begin{subfigure}{0.49\textwidth}
        \centering
        \includegraphics[width=\textwidth]{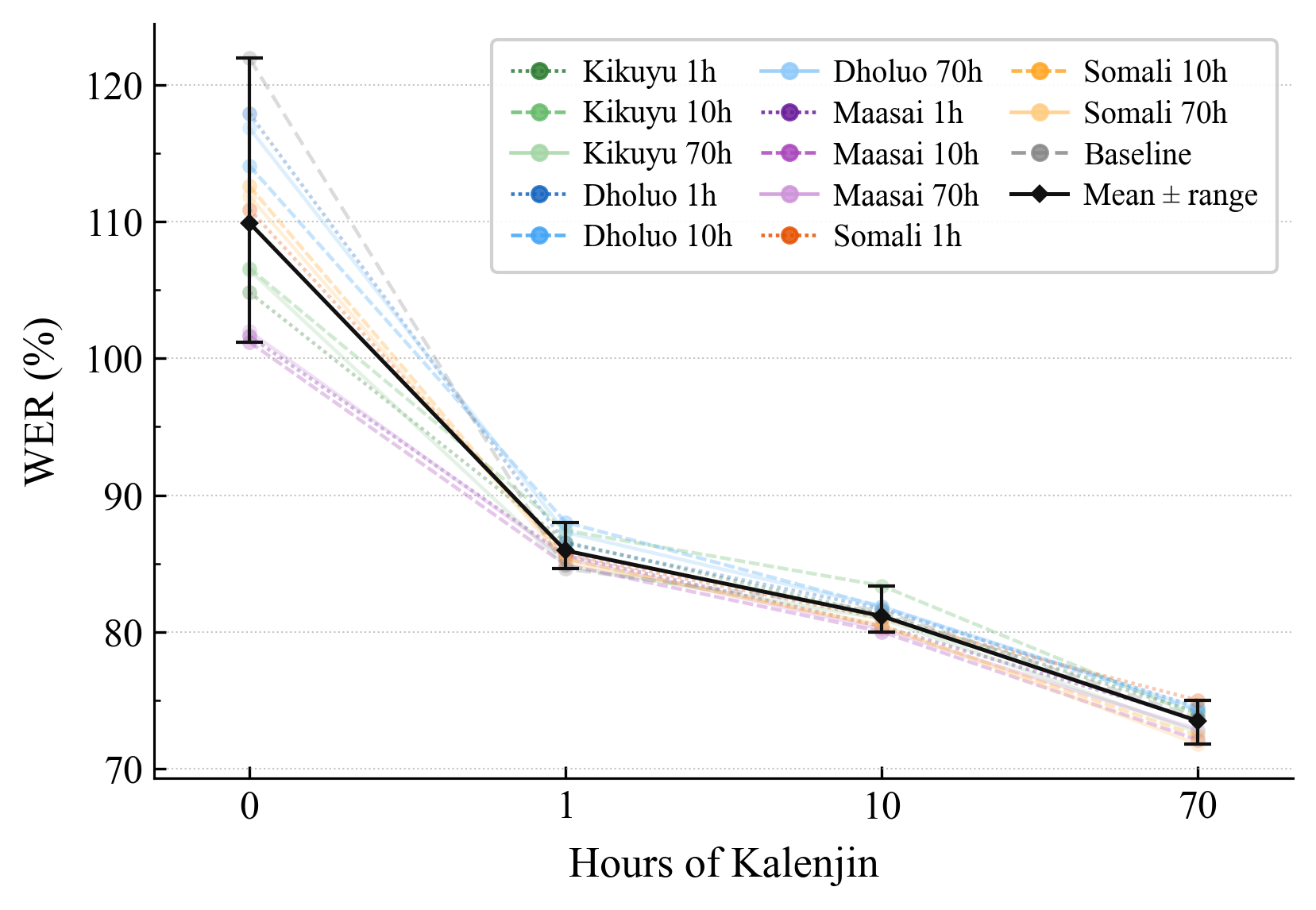}
        \label{fig:fig4a}
    \end{subfigure}
    \hfill
    \begin{subfigure}{0.49\textwidth}
        \centering
        \includegraphics[width=\textwidth]{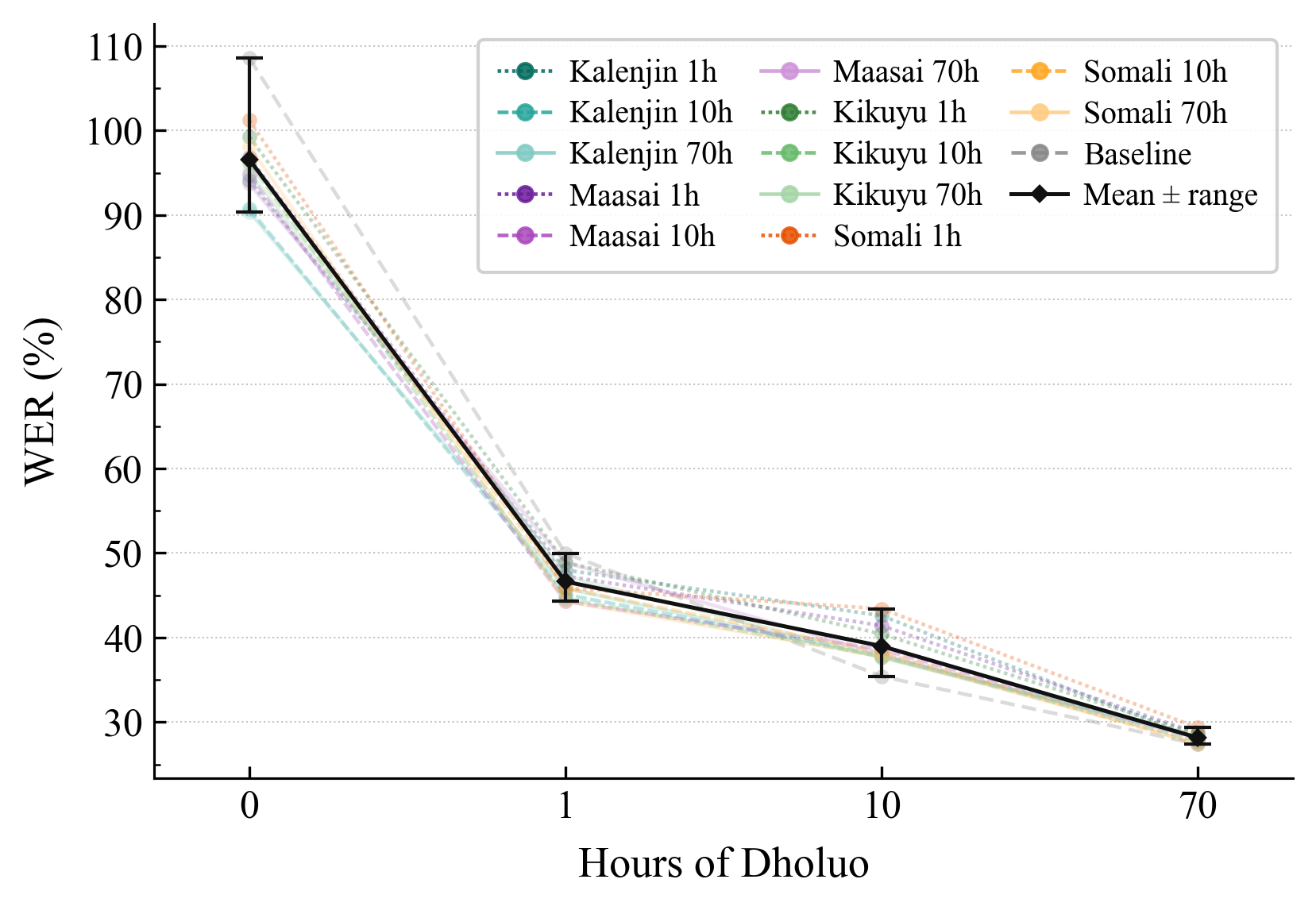}
        \label{fig:fig4b}
    \end{subfigure}
    \begin{subfigure}{0.49\textwidth}
        \centering
        \includegraphics[width=\textwidth]{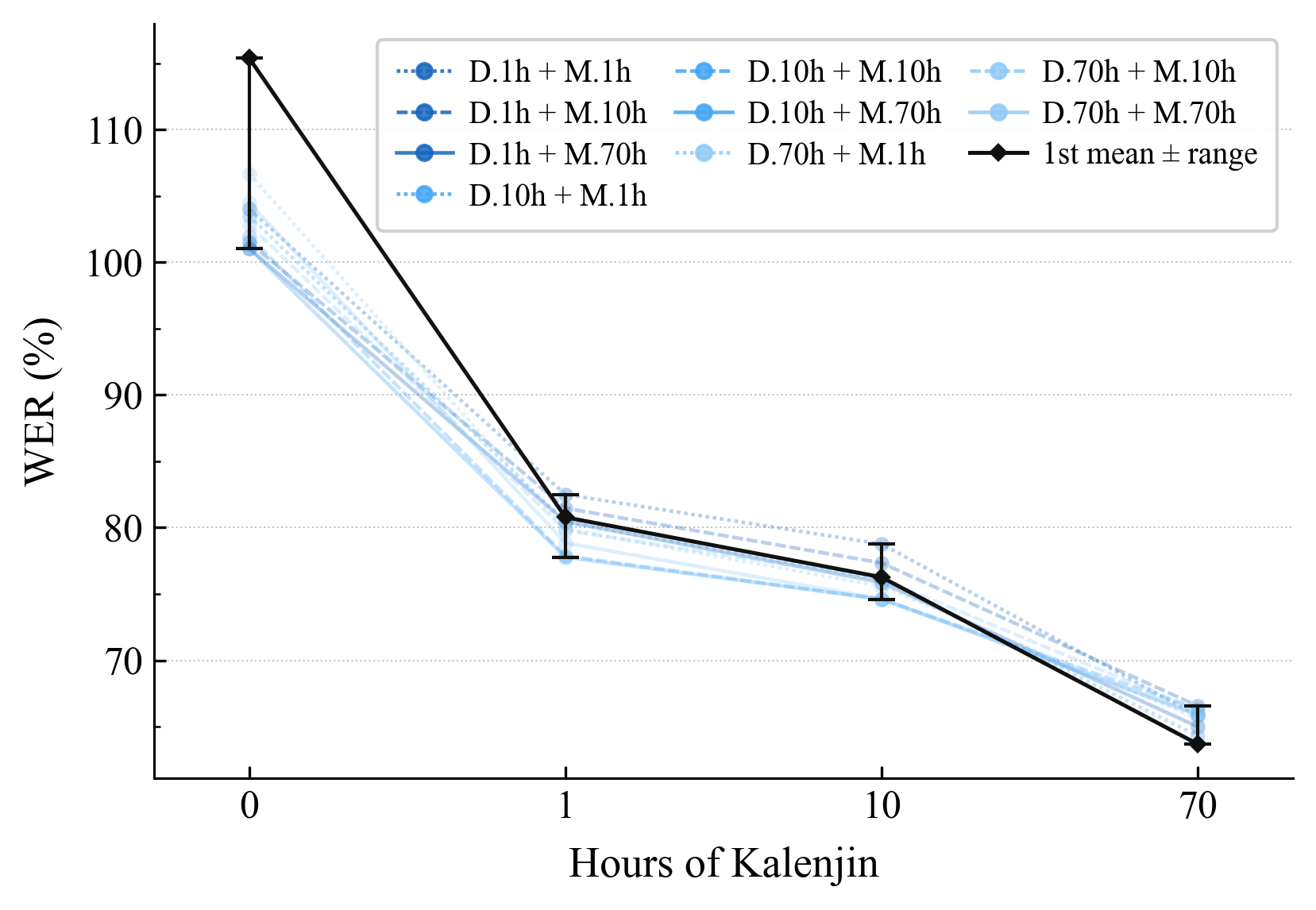}
        \label{fig:fig4c}
    \end{subfigure}
    \hfill
    \begin{subfigure}{0.49\textwidth}
        \centering
        \includegraphics[width=\textwidth]{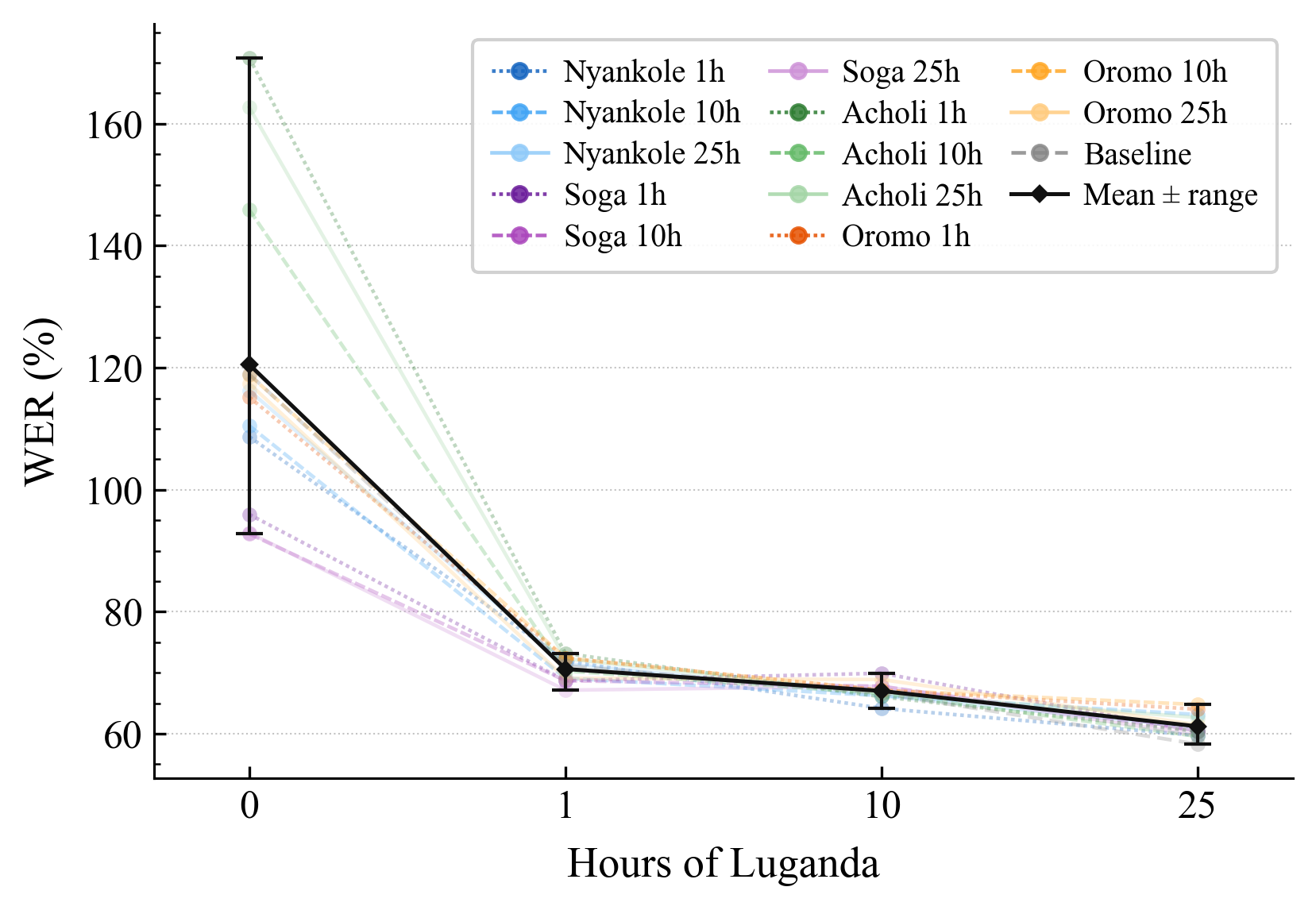}
        \label{fig:fig4d}
    \end{subfigure}
    
    \caption{WER on the target-language test set as a function of target-language fine-tuning hours across factors two through five. \textit{Top left}: factor two, replicating the first-factor setup with LoRA fine-tuning. \textit{Top right}: factor three, with Dholuo as the target language and Kalenjin reassigned to the auxiliary set. \textit{Bottom left}: factor four, in which the model is pre-adapted sequentially on Dholuo and then Maasai prior to Kalenjin fine-tuning; in this plot, the black reference line denotes the cross-model mean from factor one. \textit{Bottom right}: factor five, using the WaxalNLP corpus with Luganda as the target language and Nyankole, Soga, Acholi, and Oromo as auxiliaries. Colours denote language family and the cross-model mean and range across auxiliary data volumes are marked in black.
}
    \label{fig:fig4}
\end{figure*}


In the first factor, WER on the Kalenjin test set decreases monotonically with additional hours of target-language fine-tuning across all auxiliary models and the unmodified baseline, as shown in the left panel of Figure~\ref{fig:fig3}, consistent with expected adaptation rates for Whisper Small \citep{radford2022robust}. Prior to second-stage adaptation, models pre-adapted on an auxiliary language exhibit, on average, lower Kalenjin WER than the baseline, as shown in the right panel of Figure~\ref{fig:fig3}, a practically meaningful pre-adaptation advantage. The related auxiliaries hold a practically meaningful advantage over the baseline yet are practically equivalent to the unrelated auxiliaries, indicating this advantage stems from pre-adaptation in general, not relatedness. From one hour of Kalenjin fine-tuning onward, models pre-adapted on a related language, an unrelated language, and the Whisper Small baseline converge to a narrow band, with performance practically equivalent across all groups.


\begin{figure}[t!]
    \centering
    \includegraphics[width=\linewidth]{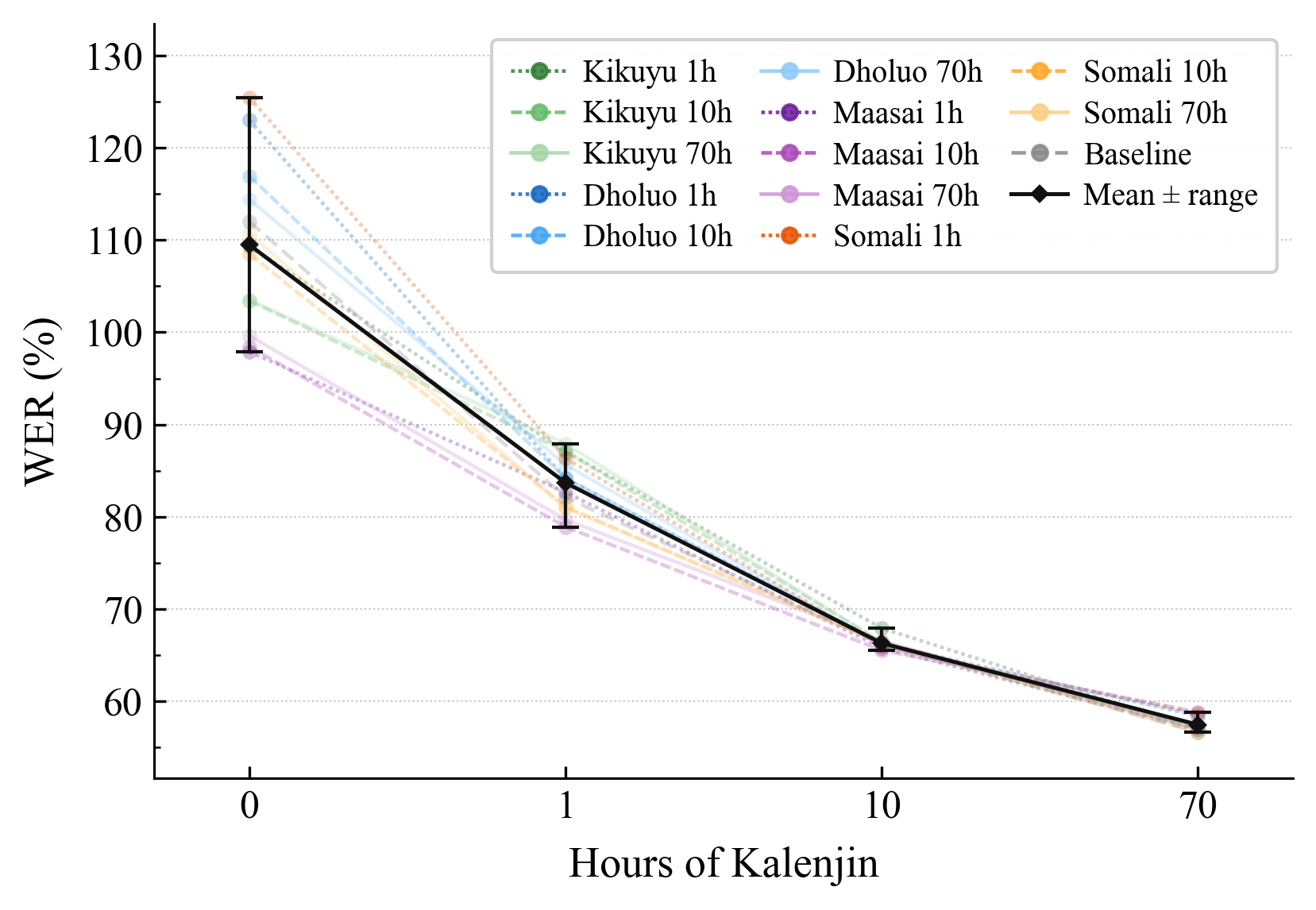}
    \vspace{0.5em}
    \includegraphics[width=\linewidth]{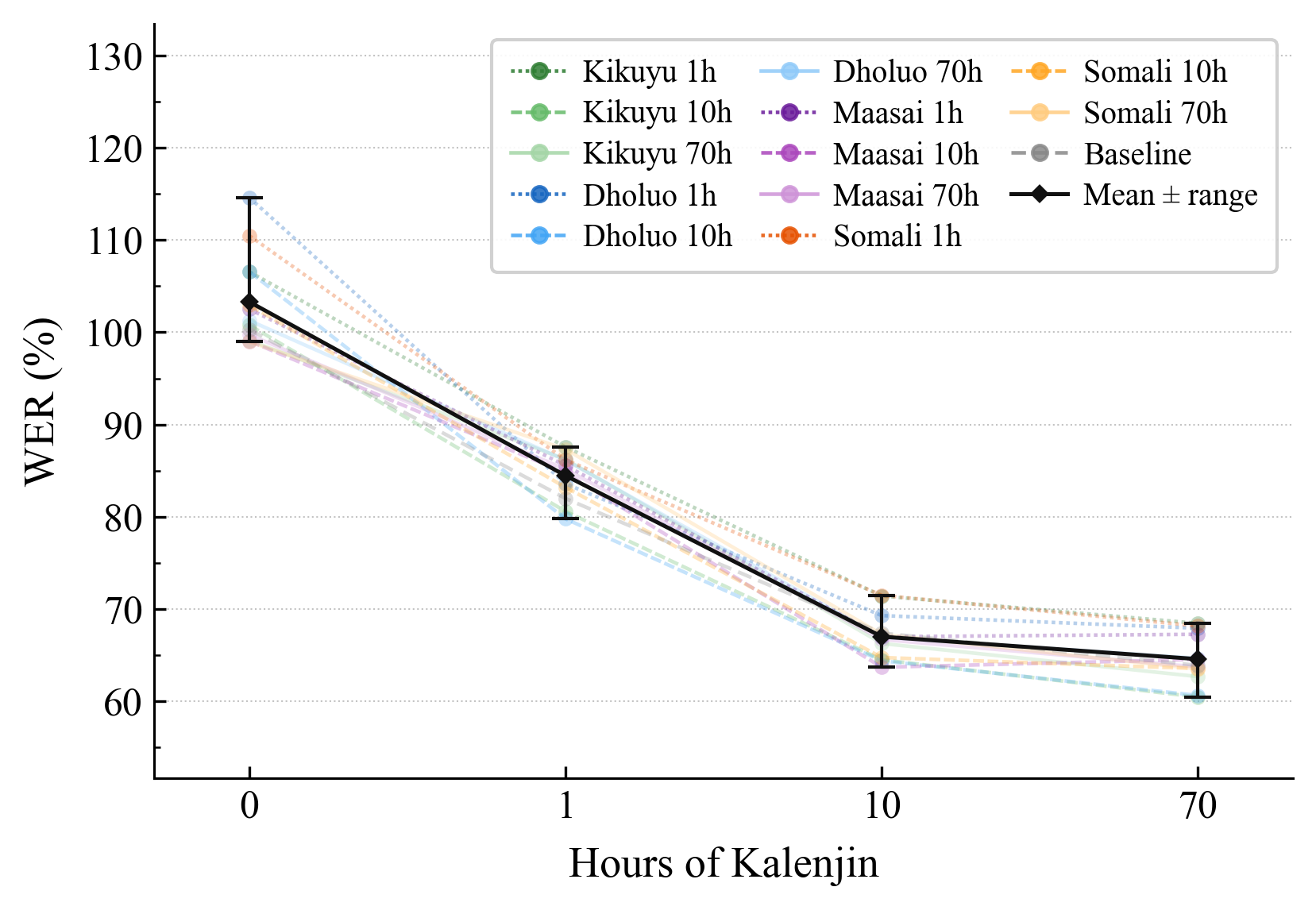}
    \vspace{0.5em}
    \includegraphics[width=\linewidth]{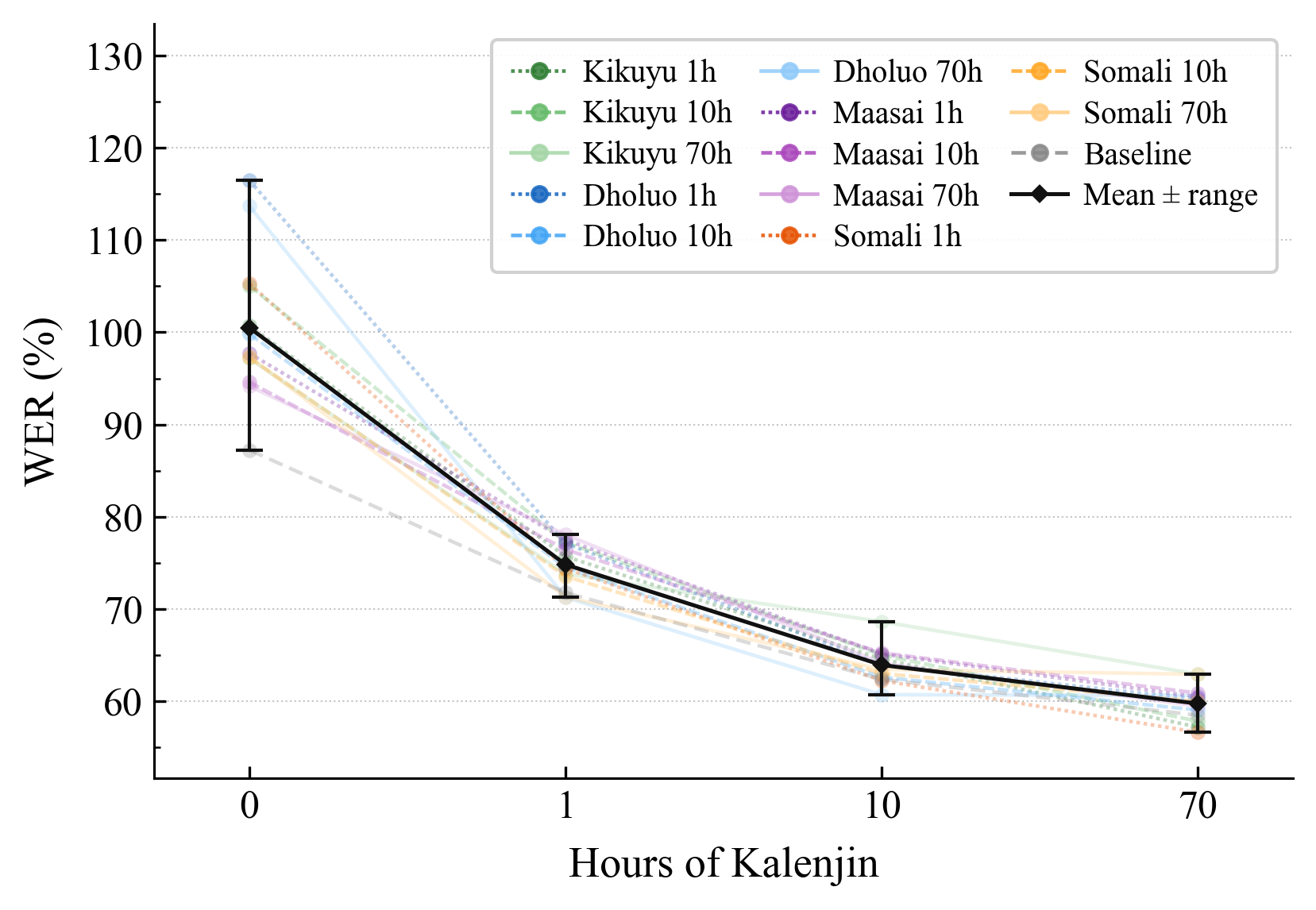}
    \caption{WER on the Kalenjin test set as a function of fine-tuning hours across all auxiliary conditions for Whisper Large v3 (top), XLS-R (centre), and OmniASR (bottom). Colours denote language family. Cross-model mean and range across auxiliary data volumes in black.}
    \label{fig:fig5}
\end{figure}


Replacing full-model fine-tuning with parameter-efficient LoRA across both fine-tuning stages produces qualitatively identical results, shown in the top-left panel of Figure~\ref{fig:fig4}. Prior to second-stage adaptation, the auxiliary models exhibit greater variance in Kalenjin WER under LoRA than under full fine-tuning. Nonetheless, differences among auxiliary groups prior to Kalenjin fine-tuning are practically equivalent regardless of relatedness, though related auxiliary models again retain a practically meaningful advantage over the unmodified baseline. Similarly to the first factor, all models converge to a narrow performance band beginning at one hour of Kalenjin fine-tuning, with target performance practically equivalent across groups.

Exchanging the roles of Dholuo and Kalenjin --- making Dholuo the target and Kalenjin an additional auxiliary --- produces the same pattern from the first factor, as shown in the top-right panel of Figure~\ref{fig:fig4}. Prior to second-stage adaptation, related auxiliary models show a practically meaningful advantage over the baseline but are practically equivalent to the unrelated auxiliaries. From one hour of Dholuo fine-tuning onward, performance is practically equivalent across all groups.

Pre-adapting sequentially on two linguistically related auxiliary languages --- Dholuo followed by Maasai, both Nilotic --- yields a practically meaningful WER advantage on the Kalenjin test set relative to both the Factor 1 unrelated auxiliaries and baseline prior to second-stage adaptation, as shown in the bottom-left panel of Figure~\ref{fig:fig4}. From one hour of Kalenjin fine-tuning onward, however, performance is practically equivalent across all groups despite the additional related pre-adaptation.

In the fifth factor, using the WaxalNLP corpus with Luganda as the target language, models pre-adapted on the linguistically related Bantu auxiliaries (Nyankole and Soga) show a practically meaningful advantage over both the unrelated auxiliaries (Acholi and Oromo) and the baseline prior to second-stage adaptation, as shown in the bottom-right panel of Figure~\ref{fig:fig4}. From one hour of Luganda fine-tuning onward, performance is practically equivalent across all groups.

The same pattern generally holds across all three additional model architectures --- Whisper Large v3, XLS-R, and OmniASR --- as shown in Figure~\ref{fig:fig5}. Prior to second-stage adaptation, OmniASR shows a practically meaningful disadvantage relative to the baseline, while Whisper Large v3 and XLS-R show no practically meaningful difference. Across all models, performance is practically equivalent from one hour of Kalenjin fine-tuning onward, a pattern that holds despite differences in prior exposure to the experiment languages across models.

Taken together, across all six experimental factors, all auxiliary and target data volumes, and all ASR models tested, pre-adaptation on a related auxiliary language yields no practically meaningful improvement in target-language performance once second-stage adaptation begins. One hour of target-language fine-tuning is sufficient to establish practical equivalence among models with related, unrelated, or no pre-adaptation across all factors (complete per-condition WER values in Table~\ref{tab:wer_all_factors}).


\section{Conclusion}

Our work evaluates whether linguistic similarity across genealogical, phonological, morphological, and syntactic dimensions can be leveraged to improve cross-lingual transfer in large multilingual ASR and reduce the volume of target-language data required to extend such models to low-resource languages, focusing on the African setting. 

We adopt the two-stage sequential fine-tuning framework of \citet{pillai2024multistage}, which leverages cross-lingual representations formed by multilingual ASR models through pre-adaptation on a related auxiliary language prior to fine-tuning on the low-resource target, with the aim of improving target-language performance. We embed this methodology, which has been shown to improve cross-lingual transfer performance for a genealogically related language pair in Whisper, in a systematic controlled experimental design spanning six factors, two Africa-centric corpora, and four large ASR models of varying scales, architectures, and supervision paradigms, with the aim of isolating whether linguistic relatedness between auxiliary and target languages constitutes a reliable predictor of cross-lingual transfer gains in these settings.

Prior to target-language fine-tuning, pre-adaptation yields a practically meaningful advantage over the baseline in some cases, yet one largely indistinguishable between related and unrelated auxiliaries, reflecting pre-adaptation in general rather than a relatedness-driven advantage. Furthermore, this advantage does not persist once target adaptation begins: across all experimental conditions, we find no practically meaningful difference in target-language performance among models fine-tuned exclusively on the target language, pre-adapted on a related auxiliary language, and pre-adapted on an unrelated auxiliary language, from as little as one hour of target-language adaptation onward. Taken together, these results suggest that linguistic relatedness alone may not reliably predict cross-lingual transfer gains in large multilingual ASR once even a trivial amount of target data is available, and is thus unlikely to constitute an effective strategy for extending such models to low-resource languages.


\section*{Limitations}

Our study considers linguistic similarity across genealogical, phonological, morphological, and syntactic dimensions, which together capture several of the axes most commonly studied in relation to cross-lingual transfer \cite{lin2019choosing, eronen2023zero}, though other linguistic characteristics may also likely play a role. Relatedly, our experimental design necessarily relies on broad typological characterisation and corpus-level similarity analyses as proxies for linguistic relatedness in forming language groupings, given the limited rigorous linguistic resources available for the low-resource languages considered. More broadly, cross-lingual transfer is likely shaped by factors beyond linguistic relatedness alone --- such as audio recording conditions, speaker demographics, and domain --- that fall outside this work's purview. Absolute WER remains high across most conditions, limiting the practical utility of the adapted models.

Our conclusions assume an equivalence bound of $\Delta = 5$ WER points as the threshold for practical meaningfulness, as outlined in Section 4. Our experimental design and results are limited to four large ASR models and two Africa-centric corpora representing ten languages across three families; confidence intervals around reported WER differences are accordingly specific to the test sets of languages considered in our experiments, and do not generalise to other languages. Broader coverage of ASR models and corpora across African geographies and language families, alongside replication in other low-resource settings, would further strengthen the generalisability of our findings.


\section*{Ethics Statement and Broader Impact}

Our work supports ASR development for low-resource African languages by studying whether linguistic relatedness improves cross-lingual transfer in large multilingual ASR, with the aim of informing more principled adaptation strategies for this setting. All experiments use the publicly available AfriVoices KE \cite{wanzare2026afrivoices} and Google WaxalNLP \cite{diack2026waxal} datasets compiled with the involvement of native speaker communities. Neither dataset contains private user data and all results are reported at the language level. Both corpora are used solely for research purposes. The accompanying code is made publicly available under GPL-3.0. We do not anticipate direct harms arising from this research.


\section*{Use of AI Assistance}

Claude Code and Claude (Anthropic) were used to assist in writing and testing code for data processing, model fine-tuning, evaluation, and data analysis. All generated code was manually reviewed and verified for correctness. Claude and Elicit AI were used to support the literature review process. All scientific content, experimental design, analysis, and conclusions are entirely the authors' own.


\bibliography{custom}


\appendix


\section*{A \quad Dataset Statistics}
\label{sec:appendix_data}

Table \ref{tab:dataset_stats} reports the dataset statistics for all languages across both corpora after filtering recordings of a length greater than 30 seconds as outlined in Section~4. The data volumes sampled for auxiliary and target fine-tuning in each factor are capped at 70 hours for AfriVoices KE and 25 hours for WaxalNLP, corresponding to the smallest per-language training sets available after filtering: Maasai (75.21 hours) and Acholi (25.08 hours) respectively.


\begin{table*}[t]
    \centering
    \small
    \renewcommand{\arraystretch}{1.15}
    \begin{tabular*}{\textwidth}{@{\extracolsep{\fill}}llcrrr@{}}
    \toprule
    \textbf{Corpus} & \textbf{Language} & \textbf{Family} &
    \textbf{Train (h)} & \textbf{Test (h)} & \textbf{Dev (h)} \\
    \midrule
    \multirow{5}{*}{AfriVoices KE}
        & Kalenjin & Nilotic  & 127.75 & 14.91 &  7.48 \\
        & Dholuo   & Nilotic  & 123.74 & 14.45 &  7.23 \\
        & Maasai   & Nilotic  &  75.21 &  8.85 &  4.42 \\
        & Kikuyu   & Bantu    & 104.15 & 12.34 &  6.21 \\
        & Somali   & Cushitic &  92.15 & 10.88 &  5.48 \\
    \midrule
    \multirow{5}{*}{Google WaxalNLP}
        & Luganda  & Bantu    &  26.11 &  3.05 &  1.55 \\
        & Nyankole & Bantu    &  37.53 &  4.36 &  2.20 \\
        & Soga     & Bantu    &  33.17 &  3.86 &  1.93 \\
        & Acholi   & Nilotic  &  25.08 &  2.94 &  1.48 \\
        & Oromo    & Cushitic & 188.68 & 22.30 & 11.17 \\
    \bottomrule
    \end{tabular*}
    \caption{Dataset statistics for all languages across both corpora after filtering, showing language family membership and train/test/dev data volumes in hours across each language.}
    \label{tab:dataset_stats}
\end{table*}


\section*{B \quad WER Results}
\label{sec:appendix_wer}

Table~\ref{tab:wer_all_factors} reports WER across the full evaluation set for all experimental conditions underlying the figures and discussion in Section~5, with auxiliary model performance averaged across all auxiliary data volumes at each target fine-tuning step for clarity, given the narrow range across auxiliary volumes at each target-data level.


\begin{table*}[t]
    \centering
    \small
    \renewcommand{\arraystretch}{1.15}
    \begin{tabular*}{\textwidth}{@{\extracolsep{\fill}}lllrrrr@{}}
    \toprule
    \textbf{Factor} & \textbf{Target} & \textbf{Auxiliary} &
    \textbf{None} & \textbf{1 hr} & \textbf{10 hrs} & \textbf{Max} \\
    \midrule
    \multirow{5}{*}{Factor 1} & \multirow{5}{*}{Kalenjin}
        & baseline  & 121.96 & 81.12 & 75.92 & 62.96 \\
    &   & luo (avg) & 122.20 & 80.48 & 77.24 & 64.38 \\
    &   & mas (avg) & 105.79 & 80.75 & 76.06 & 63.92 \\
    &   & kik (avg) & 110.13 & 80.28 & 75.57 & 63.35 \\
    &   & som (avg) & 121.22 & 81.50 & 76.30 & 63.32 \\
    \midrule
    \multirow{5}{*}{Factor 2} & \multirow{5}{*}{Kalenjin}
        & baseline  & 121.96 & 84.63 & 81.53 & 73.73 \\
    &   & luo (avg) & 116.29 & 87.26 & 81.80 & 74.39 \\
    &   & mas (avg) & 101.61 & 85.30 & 80.21 & 72.85 \\
    &   & kik (avg) & 105.97 & 86.28 & 81.87 & 73.55 \\
    &   & som (avg) & 111.79 & 85.38 & 80.64 & 73.07 \\
    \midrule
    \multirow{5}{*}{Factor 3} & \multirow{5}{*}{Dholuo}
        & baseline  & 108.56 & 49.99 & 35.41 & 27.38 \\
    &   & kln (avg) &  91.75 & 46.30 & 39.37 & 27.96 \\
    &   & mas (avg) &  95.10 & 46.82 & 39.34 & 28.32 \\
    &   & kik (avg) &  96.86 & 46.92 & 38.61 & 28.45 \\
    &   & som (avg) &  98.64 & 45.32 & 39.81 & 28.11 \\
    \midrule
    \multirow{2}{*}{Factor 4} & \multirow{2}{*}{Kalenjin}
        & baseline          & 121.96 & 81.12 & 75.92 & 62.96 \\
    &   & luo$\to$mas (avg) & 103.00 & 79.89 & 75.96 & 65.76 \\
    \midrule
    \multirow{5}{*}{Factor 5} & \multirow{5}{*}{Luganda}
        & baseline  & 118.94 & 71.20 & 66.64 & 58.29 \\
    &   & nyn (avg) & 111.82 & 70.84 & 65.55 & 61.31 \\
    &   & sog (avg) &  93.89 & 68.19 & 68.50 & 60.19 \\
    &   & ach (avg) & 159.76 & 71.92 & 66.42 & 60.88 \\
    &   & orm (avg) & 117.14 & 71.31 & 67.73 & 63.43 \\
    \midrule
    \multirow{5}{*}{Factor 6: Whisper Large} & \multirow{5}{*}{Kalenjin}
        & baseline  & 112.01 & 82.12 & 66.23 & 56.66 \\
    &   & luo (avg) & 118.12 & 84.66 & 66.16 & 57.98 \\
    &   & mas (avg) &  98.66 & 80.42 & 65.96 & 57.63 \\
    &   & kik (avg) & 105.43 & 87.36 & 66.91 & 57.05 \\
    &   & som (avg) & 114.88 & 82.81 & 66.18 & 57.47 \\
    \midrule
    \multirow{5}{*}{Factor 6: XLS-R} & \multirow{5}{*}{Kalenjin}
        & baseline  & 100.30 & 81.99 & 67.41 & 63.89 \\
    &   & luo (avg) & 107.49 & 83.17 & 66.88 & 64.41 \\
    &   & mas (avg) & 100.44 & 85.16 & 65.80 & 65.18 \\
    &   & kik (avg) & 102.16 & 84.80 & 67.40 & 63.87 \\
    &   & som (avg) & 104.13 & 85.57 & 67.84 & 65.08 \\
    \midrule
    \multirow{5}{*}{Factor 6: OmniASR} & \multirow{5}{*}{Kalenjin}
        & baseline  &  87.25 & 71.78 & 62.39 & 58.49 \\
    &   & luo (avg) & 110.01 & 74.30 & 62.43 & 59.97 \\
    &   & mas (avg) &  95.49 & 77.34 & 64.84 & 60.29 \\
    &   & kik (avg) & 100.98 & 75.58 & 66.08 & 59.33 \\
    &   & som (avg) &  99.97 & 73.10 & 62.90 & 59.86 \\
    \bottomrule
    \end{tabular*}
    \caption{Word Error Rate (WER) across target-language data volumes for all six experimental factors, reported as corpus-level WER (lower is better). \textit{None} denotes models not further adapted on target-language data: for \textit{baseline} rows this is the unmodified model, and for auxiliary rows the model fine-tuned exclusively on the listed auxiliary language. Values exceeding 100 occur where the number of words in a prediction outnumber those in the reference, reflecting the tendency of models without target-language adaptation to over-generate; such values are confined to the \textit{None} column. \textit{Max} denotes the per-corpus ceiling on target-language data (70 hours for AfriVoices KE and 25 hours for Google WaxalNLP in Factor 5). Auxiliary rows are averaged across auxiliary data volumes. In Factor 4, Whisper Small is pre-adapted sequentially on both Nilotic auxiliaries and compared against the Factor 1 baseline; its single averaged row covers all combinations of the two auxiliary data volumes. Factor 6 is reported separately for each of the three models tested: Whisper Large v3, XLS-R, and OmniASR.}    \label{tab:wer_all_factors}
\end{table*}


\section*{C \quad Statistical Analysis}
\label{sec:appendix_stats}

Tables~\ref{tab:appendix_claim_a} and~\ref{tab:appendix_claim_b} report the complete results of the two statistical analyses described in Section~4. Table~\ref{tab:appendix_claim_a} addresses whether linguistic relatedness between the auxiliary and target improves cross-lingual transfer, comparing related against unrelated auxiliary models at each volume of target-language data. Table~\ref{tab:appendix_claim_b} addresses whether auxiliary pre-adaptation improves target-language performance at all, comparing related auxiliary models against the target-only fine-tuning baseline.


\begin{table*}[h!]
\setlength{\tabcolsep}{0pt}
\renewcommand{\arraystretch}{1.1}
\small
\centering
\begin{tabular*}{\textwidth}{@{\hspace{4pt}}l@{\extracolsep{\fill}}rrr@{\hspace{4pt}}}
\toprule
\multicolumn{1}{c}{\textbf{Factor}} & \multicolumn{1}{c}{\textbf{Target (h)}} & \multicolumn{1}{c}{\textbf{$\Delta$ WER (\%)}} & \multicolumn{1}{c}{\textbf{90\% CI}} \\
\midrule
    \multirow{4}{*}{Factor 1} & 0 & $-1.68$ & $[-2.06, -1.31]$ \\
     & 1 & $-0.28$ & $[-0.41, -0.15]$ \\
     & 10 & $+0.71$ & $[+0.54, +0.89]$ \\
     & 70 & $+0.82$ & $[+0.68, +0.96]$ \\
\midrule
    \multirow{4}{*}{Factor 2} & 0 & $+0.07$ & $[-0.15, +0.29]$ \\
     & 1 & $+0.45$ & $[+0.30, +0.59]$ \\
     & 10 & $-0.25$ & $[-0.37, -0.12]$ \\
     & 70 & $+0.31$ & $[+0.18, +0.44]$ \\
\midrule
    \multirow{4}{*}{Factor 3} & 0 & $-4.33$ & $[-4.47, -4.20]$ \\
     & 1 & $+0.44$ & $[+0.29, +0.58]$ \\
     & 10 & $+0.14$ & $[-0.00, +0.29]$ \\
     & 70 & $-0.14$ & $[-0.31, +0.04]$ \\
\midrule
    \multirow{4}{*}{Factor 4} & $\bm{0}$ & $\bm{-12.67}$ & $\bm{[-13.06, -12.29]}$ \\
     & 1 & $-1.00$ & $[-1.17, -0.83]$ \\
     & 10 & $+0.02$ & $[-0.19, +0.24]$ \\
     & 70 & $+2.42$ & $[+2.19, +2.66]$ \\
\midrule
    \multirow{4}{*}{Factor 5} & $\bm{0}$ & $\bm{-35.59}$ & $\bm{[-38.26, -33.07]}$ \\
     & 1 & $-2.12$ & $[-3.17, -1.08]$ \\
     & 10 & $-0.05$ & $[-1.23, +1.17]$ \\
     & 25 & $-1.40$ & $[-2.52, -0.28]$ \\
\midrule
    \multirow{4}{*}{Factor 6: Whisper Large} & 0 & $-1.76$ & $[-1.99, -1.53]$ \\
     & 1 & $-2.55$ & $[-2.74, -2.35]$ \\
     & 10 & $-0.48$ & $[-0.64, -0.31]$ \\
     & 70 & $+0.55$ & $[+0.40, +0.69]$ \\
\midrule
    \multirow{4}{*}{Factor 6: XLS-R} & 0 & $+0.81$ & $[+0.66, +0.96]$ \\
     & 1 & $-1.02$ & $[-1.14, -0.90]$ \\
     & 10 & $-1.28$ & $[-1.40, -1.15]$ \\
     & 70 & $+0.33$ & $[+0.20, +0.45]$ \\
\midrule
    \multirow{4}{*}{Factor 6: OmniASR} & 0 & $+2.28$ & $[+2.08, +2.47]$ \\
     & 1 & $+1.48$ & $[+1.33, +1.64]$ \\
     & 10 & $-0.86$ & $[-1.00, -0.72]$ \\
     & 70 & $+0.54$ & $[+0.41, +0.66]$ \\
\bottomrule
\end{tabular*}
\caption{Per-corpus comparison of models pre-adapted on a linguistically related auxiliary language against those pre-adapted on an unrelated auxiliary, at each volume of target-language data. Each $\Delta$ is the length-weighted mean WER difference between the group of models pre-adapted on related languages and the group of models pre-adapted on unrelated languages, where a negative value favours the related auxiliary. Bracketed intervals are $90\%$ confidence intervals obtained by a non-parametric bootstrap over test utterances ($5{,}000$ resamples). Practical significance is assessed against a symmetric equivalence bound of $\Delta = 5$ WER percentage points \citep{lakens2017equivalence}: a difference where the confidence interval lies entirely within $\pm 5$ is treated as practically equivalent and one whose interval lies wholly beyond a bound is practically meaningful (shown in \textbf{bold}). For Factor~4, models sequentially pre-adapted on Dholuo and Maasai are compared against the grouping of Kikuyu and Somali single-auxiliary models from Factor~1 on the shared Kalenjin test set.}

\label{tab:appendix_claim_a}
\end{table*}


\begin{table*}[h!]
\setlength{\tabcolsep}{0pt}
\renewcommand{\arraystretch}{1.1}
\small
\centering
\begin{tabular*}{\textwidth}{@{\hspace{4pt}}l@{\extracolsep{\fill}}rrr@{\hspace{4pt}}}
\toprule
\multicolumn{1}{c}{\textbf{Factor}} & \multicolumn{1}{c}{\textbf{Target (h)}} & \multicolumn{1}{c}{\textbf{$\Delta$ WER (\%)}} & \multicolumn{1}{c}{\textbf{90\% CI}} \\
\midrule
    \multirow{4}{*}{Factor 1} & $\bm{0}$ & $\bm{-7.96}$ & $\bm{[-8.70, -7.21]}$ \\
     & 1 & $-0.50$ & $[-0.76, -0.24]$ \\
     & 10 & $+0.73$ & $[+0.43, +1.01]$ \\
     & 70 & $+1.19$ & $[+0.96, +1.42]$ \\
\midrule
    \multirow{4}{*}{Factor 2} & $\bm{0}$ & $\bm{-13.01}$ & $\bm{[-13.74, -12.31]}$ \\
     & 1 & $+1.65$ & $[+1.43, +1.88]$ \\
     & 10 & $-0.53$ & $[-0.72, -0.33]$ \\
     & 70 & $-0.11$ & $[-0.34, +0.12]$ \\
\midrule
    \multirow{4}{*}{Factor 3} & $\bm{0}$ & $\bm{-15.14}$ & $\bm{[-15.54, -14.74]}$ \\
     & 1 & $-3.43$ & $[-3.72, -3.13]$ \\
     & 10 & $+3.95$ & $[+3.67, +4.21]$ \\
     & 70 & $+0.76$ & $[+0.50, +1.02]$ \\
\midrule
    \multirow{4}{*}{Factor 4} & $\bm{0}$ & $\bm{-18.95}$ & $\bm{[-19.68, -18.27]}$ \\
     & 1 & $-1.22$ & $[-1.50, -0.94]$ \\
     & 10 & $+0.04$ & $[-0.32, +0.37]$ \\
     & 70 & $+2.80$ & $[+2.50, +3.09]$ \\
\midrule
    \multirow{4}{*}{Factor 5} & $\bm{0}$ & $\bm{-16.09}$ & $\bm{[-19.01, -13.41]}$ \\
     & 1 & $-1.74$ & $[-3.58, +0.12]$ \\
     & 10 & $+0.40$ & $[-1.43, +2.19]$ \\
     & 25 & $+2.46$ & $[+0.63, +4.31]$ \\
\midrule
    \multirow{4}{*}{Factor 6: Whisper Large} & 0 & $-3.62$ & $[-4.18, -3.07]$ \\
     & 1 & $+0.42$ & $[+0.12, +0.72]$ \\
     & 10 & $-0.17$ & $[-0.46, +0.13]$ \\
     & 70 & $+1.14$ & $[+0.89, +1.40]$ \\
\midrule
    \multirow{4}{*}{Factor 6: XLS-R} & 0 & $+3.65$ & $[+3.41, +3.89]$ \\
     & 1 & $+2.18$ & $[+1.95, +2.40]$ \\
     & 10 & $-1.06$ & $[-1.34, -0.77]$ \\
     & 70 & $+0.91$ & $[+0.64, +1.17]$ \\
\midrule
    \multirow{4}{*}{Factor 6: OmniASR} & $\bm{0}$ & $\bm{+15.50}$ & $\bm{[+15.12, +15.89]}$ \\
     & 1 & $+4.04$ & $[+3.79, +4.29]$ \\
     & 10 & $+1.25$ & $[+1.01, +1.50]$ \\
     & 70 & $+1.64$ & $[+1.41, +1.86]$ \\
\bottomrule
\end{tabular*}
\caption{Per-corpus comparison of models pre-adapted on a linguistically related auxiliary language against the target-only baseline, at each volume of target-language data. Each $\Delta$ is the length-weighted mean WER difference between the group of models pre-adapted on related languages and the baseline (the unmodified model at 0 hours, and the model fine-tuned exclusively on the target language at all non-zero volumes). A negative value favours the related auxiliary. Interval estimation, the equivalence bound of $\Delta = 5$ WER percentage points, and notation (\textbf{bold} for practically meaningful differences) follow Table~\ref{tab:appendix_claim_a}. For Factor~4, dual-Nilotic pre-adaptation is compared against the Factor~1 baseline on the shared Kalenjin test set.}
\label{tab:appendix_claim_b}
\end{table*}


\end{document}

%% file: defns.tex
\definecolor{tablecolor}{rgb}{0.8,0.8,0.8}
